# PK-YOLO: Pretrained Knowledge Guided YOLO for Brain Tumor Detection in Multiplanar MRI Slices

Ming Kang, Fung Fung Ting, Raphaël C.-W. Phan, Chee-Ming Ting
Monash University
Malaysia Campus
ting.cheeming@monash.edu

## Abstract

*Brain tumor detection in multiplane Magnetic Resonance Imaging (MRI) slices is a challenging task due to the various appearances and relationships in the structure of the multiplane images. In this paper, we propose a new You Only Look Once (YOLO)-based detection model that incorporates Pretrained Knowledge (PK), called PK-YOLO, to improve the performance for brain tumor detection in multiplane MRI slices. To our best knowledge, PK-YOLO is the first pretrained knowledge guided YOLO-based object detector. The main components of the new method are a pretrained pure lightweight convolutional neural network-based backbone via sparse masked modeling, a YOLO architecture with the pretrained backbone, and a regression loss function for improving small object detection. The pretrained backbone allows for feature transferability of object queries on individual plane MRI slices into the model encoders, and the learned domain knowledge base can improve in-domain detection. The improved loss function can further boost detection performance on small-size brain tumors in multiplanar two-dimensional MRI slices. Experimental results show that the proposed PK-YOLO achieves competitive performance on the multiplanar MRI brain tumor detection datasets compared to state-of-the-art YOLO-like and DETR-like object detectors. The code is available at https://github.com/mkang315/PK-YOLO.*

## 1. Introduction

Magnetic Resonance Imaging (MRI) of the brain to diagnose tumors by evaluating anatomical plane information extracted from multiplanar three-dimensional (3D) brain scans of the patient, such as axial, coronal, and sagittal planes that demonstrate different appearances and relationships of brain structures. Fig. 1 illustrates the three planes of the brain with anatomical position and direction. The angle of axial images is a cross-sectional view and is vertical to the bed of the MRI equipment, which divides the brain into superior and inferior. In contrast, the coronal plane is a frontal view of the human body, which divides the brain into anterior and posterior. The sagittal plane is a vertical view that longitudinally through the body and divides the brain into left and right sections. Within the X-Y-Z coordinate plane, axial pertains to a plane situated along the X-Y axis, aligned parallel to the ground, extending from the dorsal from the ventral. A coronal plane lies along the X-Z axis, distinguishing the rostral from the caudal. Sagittal planes are aligned along the Y-Z axis, dividing the brain into left and right sections. Images of each anatomical plane from various angles can be extracted from 3D MRI head scans. The greatest benefit of MRI multiplane imaging is its ability to perform whole-body imaging without needing to move the patients, allowing for flexible and detailed visualization of brain structures, which the other medical imaging techniques do not possess.

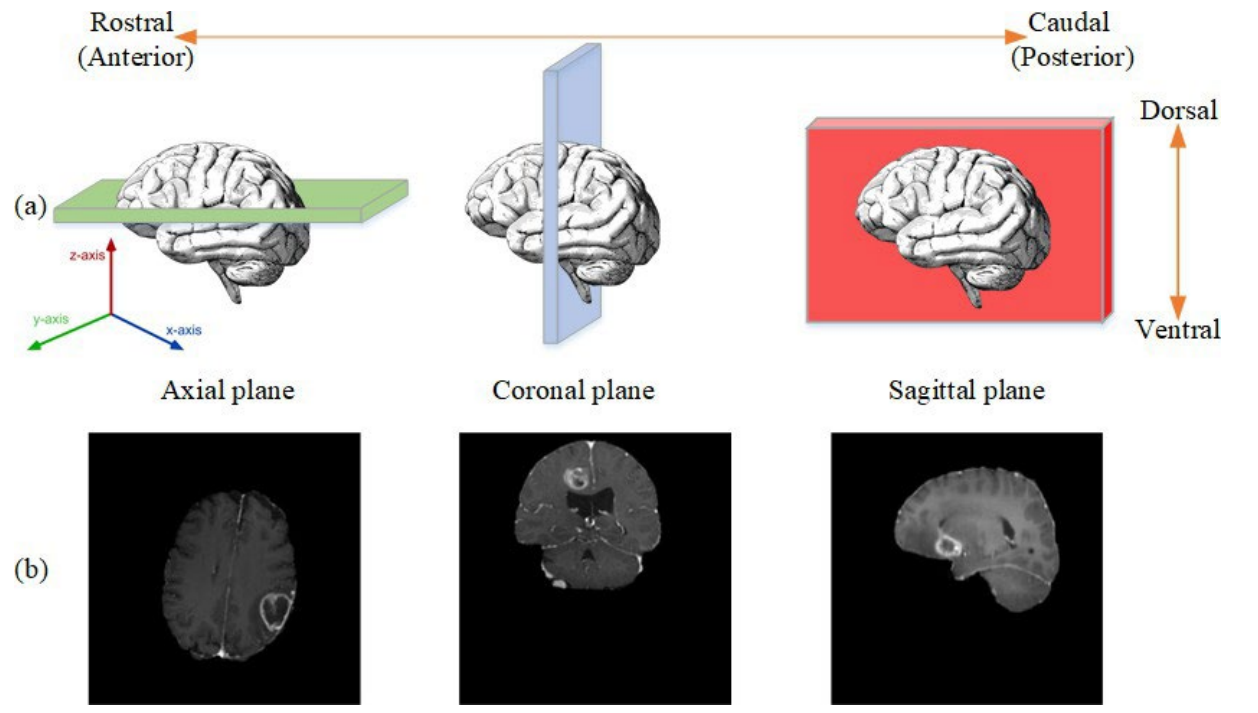

Figure 1. Examples of three planes of MRI brain scans. (a) Planes of brain sections. The brain MRI scans with a 3D structure are cut on any of the X-Y-Z planes that are named the axial, horizontal, or transverse plane (green), the coronal or frontal plane (blue), and the sagittal plane (red). Orange arrows refer to the axis of the brain. (b) Unpaired MRI slice examples of brain scans with tumors. The 3D MRI volume is divided into orthogonal slices on the classic acquisition planes: axial, coronal, and sagittal. The two-dimensional (2D) MRI slices in sectional planes are randomly selected from the Brain Tumor Object Detection Datasets [41] used in the experiments.

Although multiplanar 3D MRIs are useful for adjusting the tilt of image planes to conduct volumetric assessments of lesion volumes, 2D MRI slices have valuable advantages for diagnosis, treatment, and preoperative evaluation [24]. Each slice of a multiplanar MRI scan generates detailed images of the symptom's structures, allowing healthcare professionals to assess its anatomy and detect abnormalities [9] quickly. The advantages of multiplane imaging are that it is particularly useful in measuring the lesion size, identifying the relationship between depth and size of mass lesions relative to diagnosis, reducing visualization of artifacts in surgery, and so on. However, accurately automated lesion detection on multiplanar MRI slices is still challenging for deep learning methods due to the various lesion appearances, sizes, and relationships of internal structures in the multiplane images from different angles [36]. The number of small-size lesions in the images generally increases after converting 2D slices from 3D multiplanar MRI scans due to diverse angles of view. Particularly, it is difficult to improve the performance on all the plane images simultaneously. We investigate sorts of object detectors' performance with You Only Look Once (YOLO) and DEtection TRansformer (DETR) [3] frameworks for multiplanar MRI.

Recently, YOLOv9 [50] achieves better performances than YOLOv5 [21], YOLOv8 [23], YOLOv10 [46], and Mamba YOLO [52] on the object detection benchmark MS COCO 2017 [28] validation dataset. YOLOv9 is mainly composed of backbone, neck, and head parts, as all the typical YOLO frameworks. The backbone network effectively extracts features from input images through deep convolution operations. To tackle the issue of information loss during the layer-by-layer feature extraction and spatial transformation in deep neural networks, YOLOv9 employs two asymmetrical branches of the neck part inspired by Composite Backbone Network (CBNet) [27, 33] and DynamicDet [30], called programmable gradient information, acting as an auxiliary reversible architecture alongside the main inference branch (i.e., the original neck and head parts), which supports the main inference branch by generating reliable gradients that are then supplied to the main branch for effective backward transmission. Hence, CBFuse and CBLinear in the auxiliary branch are used as composite connections to control the main branch learning plannable multi-level semantic information. The auxiliary branch can be removed while inference. The Generalized Efficient Layer Aggregation Networks (GELAN), which is upgraded from ELAN [49] and extended ELAN [47], combines ELAN and Cross Stage Partial Network (CSPNet) [48]. The GELAN modules ensure the consistency of input and output feature sizes during feature extraction, maintaining reasonable change in the number of channels for the backbone part and optimizing parameter utilization and computational efficiency using reparameterization as computational blocks for both backbone and neck parts. Spatial Pyramid Pooling ELAN (SPPELAN), which is a connection with the backbone and neck parts, combines SPP [17] and ELAN to process multiscale images with efficient gradient propagation paths. In the head part, YOLOv9 employs the Complete Intersection over Union (CIoU) loss function [59] for bounding box regression to ensure that all aspects of the bounding box alignment are improved simultaneously.

In this paper, we explore advanced backbone networks and regression loss function improvement to further enhance object detection performance, such as RepViT [45] and Focaler-IoU [56], which can better extract robust and generalizable features and perform well in complex scenes with multiple objects. We propose a novel YOLO framework with the pretraining pipeline called PK-YOLO to improve performance for brain tumor detection in multiplane MRI slices by leveraging pretrained knowledge of RepViT backbone and Focaler-IoU loss for improving small object detection in PK-YOLO. To the best of our knowledge, this is the first pretrained knowledge guided YOLO-based object detector. The contributions of this work are summarized as follows:

1) We propose a novel pretrained Convolutional Neural Network (CNN) backbone using RepViT with Sparse masKed modeling (SparK) [42], namely SparK RepViT, to infuse domain knowledge into backbone networks, which are difficult to extract from multiplanar MRI slices for the backbone networks in the existing object detectors.

2) We propose a novel object detection pipeline for the YOLO framework by enhancing the YOLOv9 backbone network with SparK RepViT and introducing the Focaler-IoU regression loss function for our proposed PK-YOLO. The proposed modifications in PK-YOLO significantly improve tumor detection in multiplanar MRI compared to the state-of-the-art object detectors.

3) We built advanced object detectors with various pretrained backbones to compare their performance. This is the first implementation of pretrained backbones using the SparK method in YOLOs and DETRs, which includes the first end-to-end Transformer-based detector, Spark RT-DETR, for brain tumor detection.

## 2. Related work

**DETR framework.** DETR-like object detectors have obtained effective progress through advanced training schemes. Unsupervisedly Pre-train DETR (UP-DETR) [7, 8] employs an unsupervised pretrained Transformer [44] encoder and fine-tuned decoder to boost the performance of DETR with higher average precision on object detection. Group DETR v2 [5] adopts an encoder-decoder pretraining with a pretrained Vision Transformer (ViT)-Huge [12] encoder and achieves good performance on COCO test-dev. Plain-DETR [29] showcases that the backbone and decoder feature maps with masked image modeling-based backbone pretraining bring better performance for DETR. Collaborative hybrid assignments training DETR (Co-DETR) [60] with ViT-L has become the first model to achieve 66.0 average precision on COCO test-dev, but the high computational cost limits its practical application. Real-Time DETR (RT-DETR) [58], Light-Weight DETR (LW-DETR) [4], and salience DETR [19] have achieved remarkable accuracy and speed better than YOLOv5/8 and a decent performance compared to YOLOv9 on the MS COCO validation dataset. RT-DETR supports flexible adjustments of inference speed by using different decoder layers without the need for retraining, whose adaptability facilitates practical application in various real-time object detection scenarios. LW-DETR is improved with the encoder-decoder pretraining strategy. Salience DETR improves the DETR training efficiency by salience-guided supervision during training. However, there isn't related work to investigate the effectiveness of DETRs on brain tumor detection.

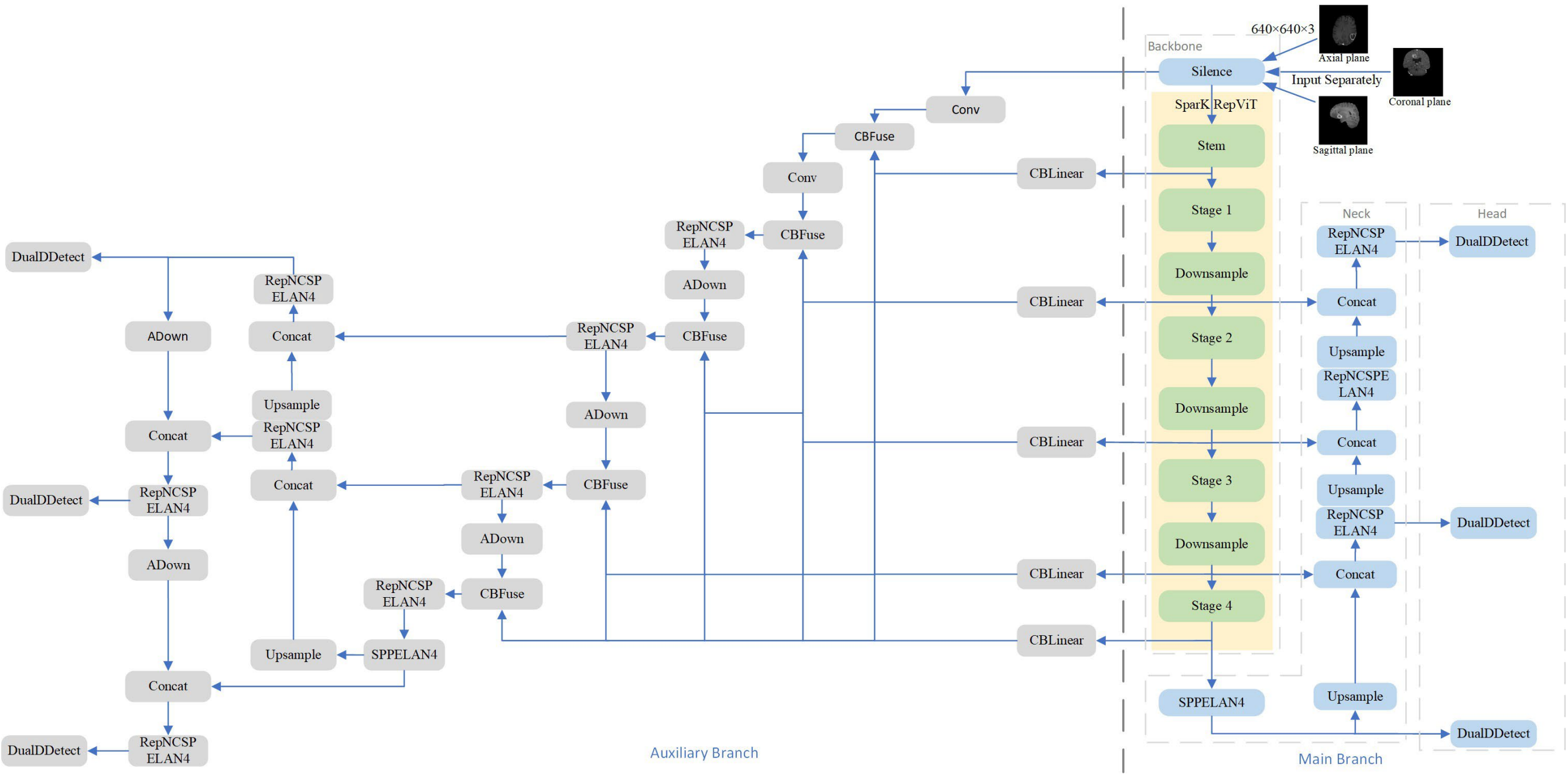

Figure 2. The overview of PK-YOLO architecture. The design of PK-YOLO is based on YOLOv9-E architecture different from YOLOv9-C's [18] and incorporates a novel SparK RepViT backbone that is highlighted with yellow background color. As an expansion of the main branch (right), there is an auxiliary branch (left) to supply reliable gradient information. CBFuse and CBLinear in the auxiliary branch are used as composite connections originating from Composite Backbone Network (CBNet). The Silence, SPPELAN4, RepNCSPELAN4, Adown, Upsample, Concat, and DualDDetect are existing modules in the source code of YOLOv9-E. Silence indicates a strategic silence operation. RepNCSPELAN4 generalizes the capability of ELAN to GELAN. ADown denotes Adaptive Downsampling. DualDDetect is a concatenation of two DDetect modules that is a fully Decoupled Detect. Three single-planar images are input separately into PK-YOLO as different sub-datasets. The details of SparK RepViT are described in Sec. 3.1 and the process of SparK pretraining is illustrated in Fig. 3.

**Brain tumor detection.** Among various tasks in medical image analysis, automated brain tumor detection, which plays an important role in medical screening, has been continuously studied in recent work. Despite YOLO's impressive performance for object detection in natural images, its applications to brain tumor detection are still limited. RCS-YOLO [25] and BGF-YOLO [26] achieved state-of-the-art performance in the brain tumor detection task.

**Multiplanar MRI analysis.** Piantadosi *et al.* [39] applied an ensemble of three different 2D deep CNNs models on slice-by-slice segmentation of breast tissues in MRI. Liu *et al.* [32] proposed a MultiPlane and multiScale Feature-level Fusion Attention model, called MPS-FFA, for Alzheimer's disease classification and employed spatial attention blocks in a multiplane feature encoder to simultaneously capture and fuse multiple pathological features in the axial, coronal, and sagittal planes. Barbato and Menga [2] trained and tested the Brain Tumor Object Detection Datasets [41] using YOLOv5m [21] and reported low detection performance in terms of mean Average Precision (mAP).

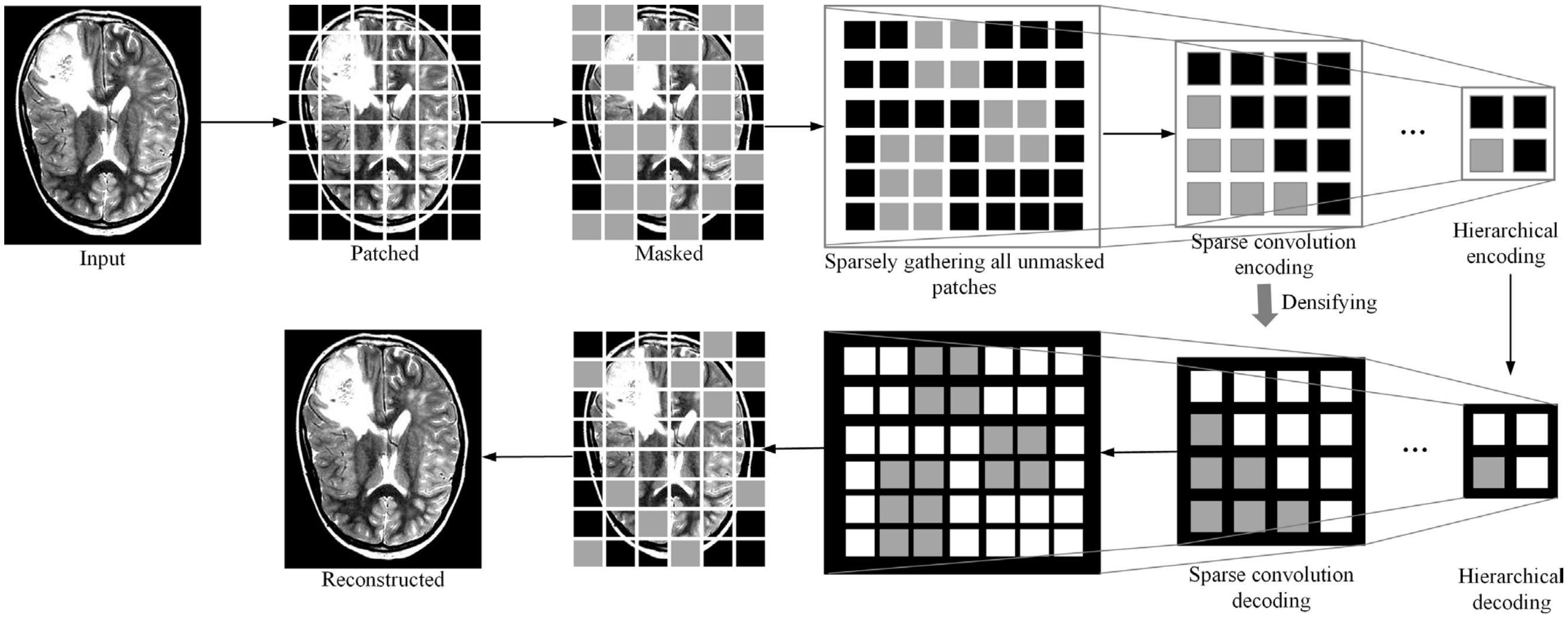

Figure 3. The process of SparK pretraining. The crux of masked image modeling lies in effectively removing pixel information from these masked patches with grey color in the figure. Sparse convolutions only compute at unmasked patches and skip masked patches whereas ordinary convolutions compute both unmasked and masked patches. We pretrain RpViT via the SparK scheme on an independent single-planar brain tumor dataset. The SparK RepViT with pretrained weights acts as the backbone of PK-YOLO shown in Fig. 2.

# 3. The proposed PK-YOLO model

In this section, we describe each component of the proposed PK-YOLO network. An overview of the proposed architecture is shown in Fig. 2. The pipeline of the proposed PK-YOLO consists of backbone pretraining and model training procedures. We first self-supervisedly pretrain a RepViT with SparK on an independent brain tumor MRI slice dataset with a single plane, then adopt the pretrained RepViT as the backbone of the YOLO architecture with improved loss, and train and test the new model with supervised learning on the multiplanar MRI slices. These enhanced methods facilitate transfer learning, allowing pretrained PK-YOLO to be adapted to detect brain tumors in the multiplanar MRI images with fewer samples and effort.

## 3.1. SparK RepViT backbone

With inspiration from the success of pretraining in DETRs, we resort to a novel pretraining pipeline in the YOLO framework. We leverage the sparse and hierarchical masked modeling technique (i.e., SparK) in the pretraining of RepViT to incorporate domain-specific knowledge that can improve the performance of the object detection model. The process of SparK pretraining of the RepViT is shown in Fig. 3. The pretrained RepViT backbone is then used to transfer domain knowledge from high-quality brain tumor datasets into the subsequent RT-DETR encoder, which guides the model to accurately detect tumors in multiplane MRI slices. We adopt SparK as the pretraining strategy due to its efficient representation and reduced computational complexity.

RepViT is a pure lightweight CNN with reparameterized depthwise convolutions [6, 11], which showcases state-of-the-art performance and optimal latency by reexamining efficient CNN designs from a ViT perspective. RepViT is composed of one stem, four stages, three deep downsample modules, one pooling layer, and one fully connected layer, as shown in Fig. 2. The stem is the early convolutions [53], including two convolutions with a stride of 2 for initial downsampling. RepViTSEBlock is inserted with a squeeze-and-excitation [20] layer between convolution and feed-forward modules in RepViTBlock. Three deep downsample layers are adopted between four adjacent stages. As depicted in Fig. 4, RepViTBlock and RepViTSEBlock are utilized for each stage in a cross-block manner and separate the token and channel mixer through structural reparameterization. This technique allows RepViT to combine the benefits of complex and simple models where the model is trained with a more complex structure and then simplified for inference, maintaining high performance while reducing complexity. The pure convolutional nature and training-inference discrepancy of RepViT offer substantial advantages in latency for various vision tasks.

Motivated by the excellent performance of masked representation learning that was proposed in Bidirectional Encoder Representations from Transformers (BERT) [10] for pretraining language models and BERT-style pretraining image models [1, 16], SparK leverage sparse and hierarchical representations to improve the efficiency and interpretability of BERT within CNN architectures. As shown in Fig. 3, the SparK pretraining process first partitions the im-age into multiple square non-overlapping patches, and each patch undergoes independent masking with a given probability. Subsequently, all unmasked patches are sparsely combined into a sparse image, and sparse convolutions of RepViT are then used to encode and decode it hierarchically. In the end, the image is reconstructed after densifying it with a sparse convolution decoder. Upon completing pretraining, the encoder is exclusively employed for the PK-YOLO detection model whereas the decoder is discarded.

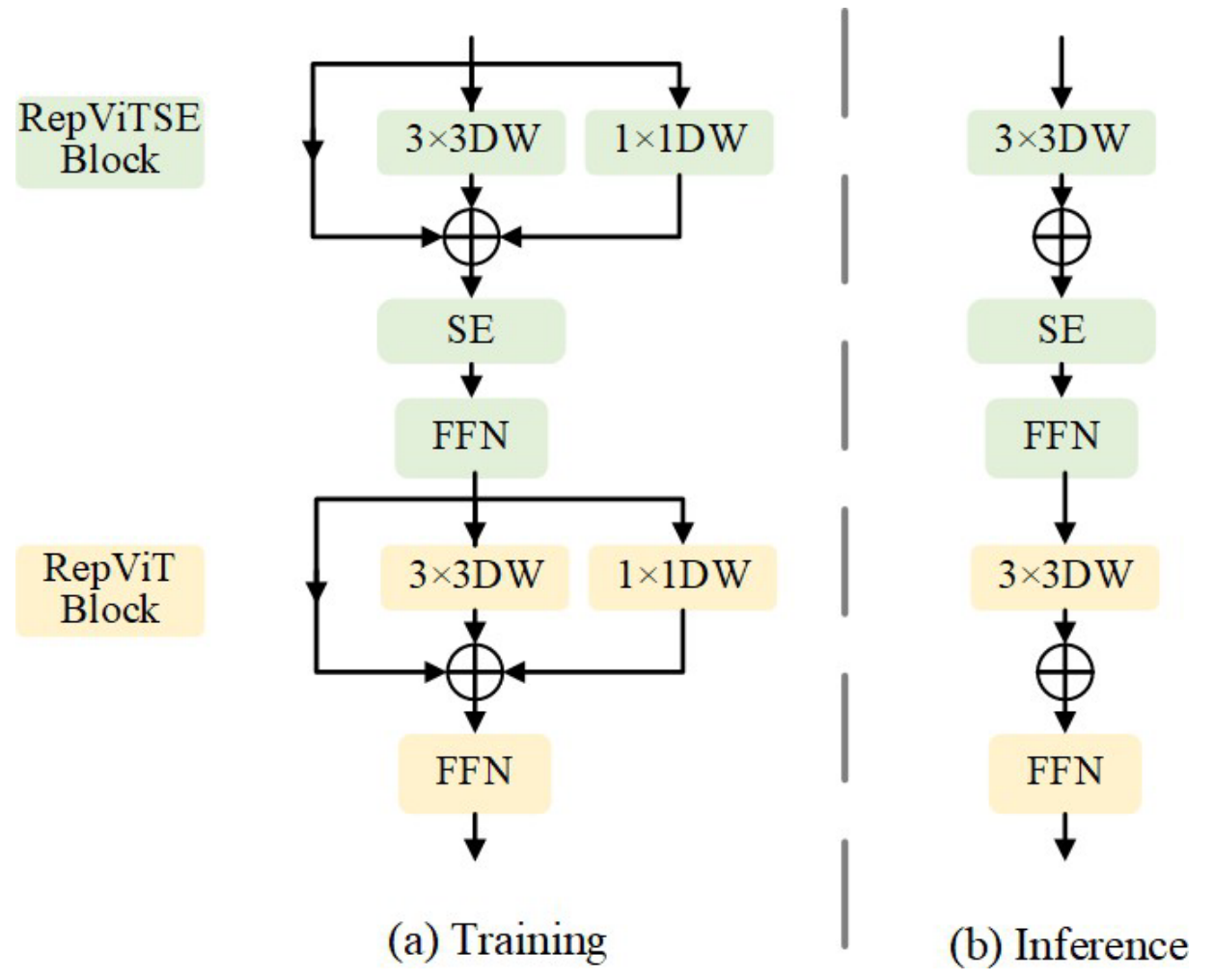

Figure 4. The structure of stage in RepViT. (a) Stage in the model training phase. (b) Stage during model inference (or deployment). There are two modules, RepViTBlock and RepViTSEBlock, in every stage, which are highlighted with green and yellow colors. RepViTBlock is composed of reparameterized Depth Wise (DW) convolution and Feed Forward Network (FFN). Squeeze-and-Excitation (SE) layer is used in the RepViTSEBlock. At the inference stage, only $3 \times 3$ convolution remains from the identity branch, $1 \times 1$ convolution and $3 \times 3$, by using structural reparameterization.

Using pretrained RepViT via SparK for the proposed model offers several advantages. First, pretrained RepViT models who are trained on high-quality brain tumor datasets, have already learned to recognize tumor features that capture various patterns and semantics knowledge of tumors in MRI. They can transfer this knowledge to object detection tasks in multiplane MRI slices. Second, pretrained RepViT can also help boost model generalization capability, which occurs when there are limited labeled data for training in detection tasks. Pretrained RepViT has learned generic features from diverse tumors of different sizes and types, enabling it to generalize well to multiplane MRI object detection tasks and datasets. Third, this customized pretrained RepViT allows for the flexible incorporation of adaptable in-domain knowledge at hand that is better suited to multiplanar MRI tumor detection.

### 3.2. Multi-level information fusion

The deep layer features during the backpropagation process in deep neural networks are closer to the output and the network parameters are easier to update, while the shallow layer features are closer to the input and more difficult to update. Therefore, the auxiliary branch can effectively provide supplementary information that is lost during the layer-by-layer transformation process for the backbone network. Different from multiple identical backbones (i.e., lead and assistant backbones) in CBNet and DynamicDet, the auxiliary branch employs multiple prediction branches that are different neck parts to carry out disparate feature fusion operations and integrate various contextual information reversibly. Multi-level gradient information from different deep feature pyramid necks guides the learning of network architectures of different sizes. Specifically, the CBLiear routing module is employed, allowing the auxiliary network to adaptively process objects of different scales to improve detection performance across small and large objects. Multi-level auxiliary information is routed by CBLiear modules from the auxiliary branch to the main branch to ensure that complete information containing various scale objects is retained, alleviating the issue of broken information in deep neural networks.

During the training process, shallow features in the auxiliary branch are directed to learn the features necessary for small object detection, while the network treats the positions of objects of other sizes as background. On the main branch side, the SparK ReViT backbone with SE layers in the proposed PK-YOLO enhances the representational power of feature maps by recalibrating channel-wise feature responses. This means that even small brain tumors can have their features amplified, making them more distinguishable from the background. Hence, our proposed model is more suited for detecting small objects after connecting with shallow layer features from the auxiliary branch, which have strong spatial information and small receptive fields. Thus, the design of asymmetrical neck branches incorporating an enhanced backbone provides an effective mechanism to obtain both primary and supplementary feature information.

### 3.3. Regression loss improvement

Focaler-IoU, a refined metric for evaluating object detection models, places greater emphasis on difficult-to-detect objects, improving the model's ability to learn from small object samples. It stabilizes the training process by reducing the influence of easily detectable objects. By adjusting the importance of different samples, this regression loss improvement in PK-YOLO helps in balancing the training process, ensuring that challenging brain tumor instances in multiplanar MRIs receive adequate attention.

| Dataset | Train | | Test | | Subtotal | |
|---|---|---|---|---|---|---|
| | Sample | Instance | Sample | Instance | Sample | Instance |
| axial_t1ce_2_class | 310 | 309 | 75 | 81 | 385 | 390 |
| coronal_t1ce_2_class | 319 | 337 | 78 | 83 | 397 | 420 |
| sagittal_t1ce_2_class | 264 | 279 | 70 | 77 | 334 | 356 |
| All | 893 | 925 | 223 | 241 | 1116 | 1166 |

Table 1. Number of samples and instances in train and test sets of evaluation datasets. The number of instances to be detected in images is more than that of samples, which means the evaluation datasets are multiple object detection datasets because there is more than one annotated brain tumor in some slices.

# 4. Experiments

## 4.1. Datasets

We pretrained the backbone network of the proposed PK-YOLO on the brain tumor dataset Br35H [14], which contains 801 high-quality MRI images in a single plane. We evaluated the performance of the proposed model on 1116 images of the public multiplane MRI slice dataset [41], which was extracted from 400 images of the T1-weighted Contrast-Enhanced (T1wCE) series in Digital Imaging and Communications in Medicine (DICOM) format of the RSNA-MICCAI Brain Tumor AI Challenge (2021) [35]. The manually annotated tumor locations and classes (i.e., negative and positive) are provided. The dataset is separated according to the three plans, i.e., axial, coronal, and sagittal, and divided into training and test sets. There are multiple annotated brain tumors in some slices, which brings difficulties in accurate detection. The number of samples and instances (i.e., annotated brain tumors) in each plane is shown in Tab. 1.

## 4.2. Implementation details

We first trained RepViT-M2.3 with SparK pretraining strategy with 300 epochs on the dataset Br35H and didn't freeze the pretrained weights, which are different from the prerequisite for the success of pretraining Transformers in DETRs [29]. Then, we trained and tested the proposed PK-YOLO on the three MRI slice datasets separately based on the initial weights of the pretrained backbone network.

We trained our proposed model PK-YOLO with NVIDIA RTX4090 GPU with 24GB memory using the stochastic gradient descent optimizer with a weight decay of 0.0005. The initial learning rate is set to 0.01 and 0.0001 for the finish. The number of training epochs on each subset was set to 300, respectively. The implementation settings of the other models in the ablation study are the same as these.

## 4.3. Results

Tab. 2 shows the performance comparison with other state-of-the-art object detectors. PK-YOLO achieves better average precision than both DETR-like and YOLO-like object detectors on all three subsets of the multiplane MRI slice dataset. Note that all the compared models are not pretrained on the brain tumor dataset. In terms of $mAP_{50}$, PK-YOLO outperforms than all the state-of-th-art DETRs, including Co-DETR with Swin L (epoch=36, DETR augmentation) and Salience DETR [19] with FocalNet-L [54]. Compared to the YOLOv9-E, our proposed PK-YOLO increases 1.2%, 7.6%, 4.8% on $mAP_{50}$, respectively. Moreover, PK-YOLO also surpasses two modified models for brain tumor detection, RCS-YOLO and BGF-YOLO. Our method achieves the highest $mAP_{50}$ for the sagittal plane; however, none of the methods can give the best performance on both $mAP_{50}$ and $mAP_{50:95}$.

A limitation of our methods is the model parameters, computation amount, and inference time are more than almost all the YOLO-like methods and parts of DETR-like methods. However, the PK-YOLO gives a substantial improvement in accuracy despite a slight increase in computational effort and is faster than Mamba YOLO-L in the inference phase.

## 4.4. Ablation studies

### 4.4.1 Impact of proposed methods

We conducted an ablation study on the role of each proposed component on the detection performance. Tab. 3 shows that the model performance declines with the omission of any of the components. RepViT backbone has the largest impact on the performance of PK-YOLO. If the RepViT is replaced by the original backbone in YOLOv9, $mAP_{50}$ of the model decreases significantly for the three planes but $mAP_{50:95}$ increases for the sagittal plane. The SparK pretraining method plays a secondary role for the axial and coronal planes. Focaler-IoU serves the same effect as RepViT only on the axial subsets.

### 4.4.2 Feature transferability of different backbones

We assess the different variants of backbone in PK-YOLO based on RT-DETR-X, YOLOv8x, YOLOv9-E, and YOLOv10-X by pretraining these backbones using the SparK technique. The backbone network of RT-DETR, the PaddlePaddle High-performance Graphics Processing Unit (GPU)-friendly Network version 2 (PP-HGNetv2 or HGNetv2) [37, 38], is a pure CNN and those of YOLOv8/9/10 are known as CNN-based backbone. Therefore, the SparK pretraining method is suitable for all the backbones in them.

| Dataset | Model | Param (M) | Precision | Recall | $mAP_{50}$ | $mAP_{50:95}$ |
|---|---|---|---|---|---|---|
| axial_t1ce_2_class | UP-DETR [7, 8] | 41.3 | – | – | 0.754 | 0.477 |
| | RT-DETR-X [58] | 65.5 | 0.662 | 0.568 | 0.538 | 0.347 |
| | LW-DETR-xlarge [4] | 118.0 | – | – | 0.926 | **0.687** |
| | Co-DETR with Swin L [60] | 218.0 | – | – | 0.893 | 0.584 |
| | Salience DETR with FocalNet-L [19] | 56.1 | – | – | 0.816 | 0.526 |
| | RCS-YOLO [25] | 45.7 | 0.944 | 0.839 | 0.839 | 0.573 |
| | BGF-YOLO [26] | 84.3 | 0.941 | 0.789 | 0.941 | 0.579 |
| | YOLOv5x [21] | 86.2 | 0.741 | 0.679 | 0.828 | 0.596 |
| | YOLOv8x [23] | 68.2 | **0.894** | 0.836 | 0.908 | 0.656 |
| | YOLOv9-E [50] | 60.5 | 0.838 | 0.877 | 0.935 | 0.667 |
| | YOLOv10-X [46] | 31.6 | 0.743 | 0.821 | 0.832 | 0.558 |
| | Mamba YOLO-L [52] | 57.6 | 0.891 | 0.753 | 0.915 | 0.666 |
| | **PK-YOLO (Ours)** | 108.7 | 0.858 | **0.896** | **0.947** | 0.681 |
| coronal_t1ce_2_class | UP-DETR | 41.3 | – | – | 0.236 | 0.186 |
| | RT-DETR-X | 65.5 | 0.742 | 0.592 | 0.575 | 0.407 |
| | LW-DETR-xlarge | 118.0 | – | – | 0.723 | 0.549 |
| | Co-DETR with Swin L | 218.0 | – | – | 0.510 | 0.304 |
| | Salience DETR with FocalNet-L | 56.1 | – | – | 0.520 | 0.343 |
| | RCS-YOLO | 45.7 | 0.493 | 0.884 | 0.574 | 0.369 |
| | BGF-YOLO | 84.3 | 0.494 | **0.889** | 0.593 | 0.417 |
| | YOLOv5x | 86.2 | 0.648 | 0.690 | 0.681 | 0.466 |
| | YOLOv8x | 68.2 | 0.672 | 0.650 | 0.697 | 0.524 |
| | YOLOv9-E | 60.5 | 0.552 | 0.788 | 0.729 | 0.526 |
| | YOLOv10-X | 31.6 | 0.519 | 0.656 | 0.605 | 0.423 |
| | Mamba YOLO-L | 57.6 | 0.641 | 0.770 | 0.758 | 0.539 |
| | **PK-YOLO (Ours)** | 108.7 | **0.834** | 0.793 | **0.805** | **0.689** |
| sagittal_t1ce_2_class | UP-DETR | 41.3 | – | – | 0.231 | 0.137 |
| | RT-DETR-X | 65.5 | 0.441 | 0.435 | 0.395 | 0.254 |
| | LW-DETR-xlarge | 118.0 | – | – | 0.471 | 0.343 |
| | Co-DETR with Swin L | 218.0 | – | – | 0.496 | 0.290 |
| | Salience DETR with FocalNet-L | 56.1 | – | – | 0.509 | 0.346 |
| | RCS-YOLO | 45.7 | 0.500 | 0.779 | 0.515 | 0.357 |
| | BGF-YOLO | 84.3 | 0.485 | 0.792 | 0.545 | 0.347 |
| | YOLOv5x | 86.2 | 0.469 | 0.840 | 0.561 | 0.370 |
| | YOLOv8x | 68.2 | 0.414 | 0.782 | 0.533 | **0.385** |
| | YOLOv9-E | 60.5 | 0.437 | **0.869** | 0.534 | 0.383 |
| | YOLOv10-X | 31.6 | 0.536 | 0.525 | 0.544 | 0.361 |
| | Mamba YOLO-L | 57.6 | **0.477** | 0.842 | 0.559 | **0.385** |
| | **PK-YOLO (Ours)** | 108.7 | 0.476 | 0.845 | **0.582** | 0.382 |

Table 2. Performance comparisons with state-of-the-art methods. Results. RT-DETR-X was implemented based on Ultralytics's code of rtdetr-x [22]. The original code of all UP-DETR, Co-DETR, LW-DETR, and Salience DETR versions only prints average precision and average recall. The best results are shown in bold.

We pretrained the convolutional backbones in the abovementioned models to obtain the pretrained backbone on the Br35H, respectively. Then we trained and tested on the same dataset as the proposed PK-YOLO. The results demonstrate feature transferability in PK-YOLO is higher than that in RT-DETR-X, YOLOv8x, YOLOv9-E, and YOLOv10-X by comparing the performance increments between improved and original models, as shown in Tab. 4. The results indicate the pretrained backbones replacing the original ones in all the models benefit performance boost.

#### 4.4.3 Effect of different loss functions

We performed an ablation study on the influence of regression losses, including CIoU [59], Generalized IoU

| Dataset | Method | Precision | Recall | $mAP_{50}$ | $mAP_{50:95}$ |
|---|---|---|---|---|---|
| axial | w/o RepViT | 0.857 | 0.814 | 0.915 | 0.655 |
| | w/o SparK | 0.905 | 0.820 | 0.935 | 0.615 |
| | w/o Focaler-IoU | 0.857 | 0.814 | 0.915 | 0.655 |
| coronal | w/o RepViT | 0.621 | 0.655 | 0.703 | 0.526 |
| | w/o SparK | 0.762 | 0.711 | 0.721 | 0.653 |
| | w/o Focaler-IoU | 0.795 | 0.755 | 0.757 | 0.687 |
| sagittal | w/o RepViT | 0.474 | 0.802 | 0.554 | 0.411 |
| | w/o SparK | 0.477 | 0.753 | 0.561 | 0.401 |
| | w/o Focaler-IoU | 0.445 | 0.872 | 0.578 | 0.414 |

Table 3. Ablation study of each method in the proposed PK-YOLO. w/o stands for without.

| Dataset | Model | Precision | Recall | $mAP_{50}$ | $mAP_{50:95}$ |
|---|---|---|---|---|---|
| axial | SparK RT-DETR-X | 0.879 | 0.519 | 0.601 | 0.414 |
| | SparK YOLOv8x | 0.839 | 0.834 | 0.920 | 0.675 |
| | SparK YOLOv9-E | 0.857 | 0.814 | 0.915 | 0.655 |
| | SparK YOLOv10-X | 0.786 | 0.765 | 0.842 | 0.563 |
| | **PK-YOLO (Ours)** | **0.858** | **0.896** | **0.947** | **0.681** |
| coronal | SparK-RT-DETR-X | 0.696 | 0.657 | 0.551 | 0.353 |
| | SparK YOLOv8x | 0.672 | 0.650 | 0.697 | 0.524 |
| | SparK YOLOv9-E | 0.621 | 0.655 | 0.703 | 0.526 |
| | SparK YOLOv10-X | 0.485 | 0.693 | 0.608 | 0.415 |
| | **PK-YOLO (Ours)** | **0.834** | **0.793** | **0.805** | **0.689** |
| sagittal | SparK RT-DETR-X | 0.575 | 0.427 | 0.448 | 0.273 |
| | SparK YOLOv8x | 0.460 | 0.717 | 0.574 | **0.427** |
| | SparK YOLOv9-E | 0.474 | 0.802 | 0.554 | 0.411 |
| | SparK YOLOv10-X | 0.423 | 0.664 | 0.532 | 0.347 |
| | **PK-YOLO (Ours)** | **0.476** | **0.845** | **0.582** | 0.382 |

Table 4. Ablation study on feature transferability of pretrained backbones. All the backbones in each model are pretrained using the same pretraining method. The best results are shown in bold.

(GIoU) [40], Normalized Wasserstein Distance (NWD) [51], $\alpha$-IoU [15], Efficient IoU (EIoU) [57], Scylla-IoU (SIoU) [13], Wise IoU (WIoU) [43], MDPIoU [34], Shape-IoU [55], Powerful-IoU [31] and Focaler-IoU [56]. The original regression loss CIoU in YOLOv5/8/9/10 has the robustness of the bounding box for object detection. As shown in Tab. 5, Focaler-IoU employed in the PK-YOLO achieves the best performance on all three subsets compared to other regression losses, expect $mAP_{50:95}$ of the sagittal plane. The $mAP_{50:95}$ on the sagittal plane can be better if GIoU replaces Focaler-IoU in PK-YOLO.

## 5. Conclusion

We developed a novel automated brain tumor detector for multiplane MRI slices by combining both a more powerful backbone and a more effective regression loss function. We desgined a pretrained SparK RepViT backbone and introduced a difficult sample focused regression loss, which enables detection of small-size tumors. The proposed PK-YOLO, with higher average precision, achieves state-of-the-art performance compared to those of the YOLOs and DETRs models on a challenging brain tumor multiplane MRI dataset. In the future work, we will evaluate PK-YOLO on the other datasets to demonstrate its generalization to different imaging domains.

| Dataset | Model | Precision | Recall | $mAP_{50}$ | $mAP_{50:95}$ |
|---|---|---|---|---|---|
| axial | CIoU [59] | 0.838 | 0.877 | 0.935 | 0.667 |
| | GIoU [40] | 0.826 | **0.940** | 0.934 | 0.647 |
| | NWD [51] | 0.954 | 0.765 | 0.921 | 0.662 |
| | $\alpha$-IoU [15] | 0.815 | 0.901 | 0.931 | 0.661 |
| | EIoU [57] | 0.801 | 0.896 | 0.907 | 0.665 |
| | SIoU [13] | 0.880 | 0.827 | 0.934 | 0.663 |
| | WIoU [43] | 0.811 | 0.903 | 0.937 | 0.644 |
| | MDPIoU [34] | 0.805 | 0.889 | 0.905 | 0.662 |
| | Shape-IoU [55] | 0.848 | 0.827 | 0.918 | 0.645 |
| | Powerful-IoU [31] | **0.941** | 0.391 | 0.815 | 0.082 |
| | **Focaler-IoU** [56] | 0.858 | 0.896 | **0.947** | **0.681** |
| coronal | CIoU | 0.552 | 0.788 | 0.729 | 0.526 |
| | GIoU | 0.764 | 0.774 | 0.765 | 0.683 |
| | NWD | 0.459 | 0.824 | 0.583 | 0.421 |
| | $\alpha$-IoU | 0.480 | 0.875 | 0.677 | 0.502 |
| | EIoU | 0.459 | 0.889 | 0.585 | 0.424 |
| | SIoU | 0.469 | 0.895 | 0.592 | 0.425 |
| | WIoU | 0.816 | 0.678 | 0.764 | 0.671 |
| | MDPIoU | 0.777 | 0.736 | 0.711 | 0.652 |
| | Shape-IoU | **0.858** | 0.772 | 0.793 | 0.709 |
| | Powerful-IoU | 0.809 | 0.777 | 0.769 | 0.682 |
| | **Focaler-IoU** | 0.834 | **0.793** | **0.805** | **0.689** |
| sagittal | CIoU | 0.437 | 0.869 | 0.534 | 0.383 |
| | GIoU | **0.497** | 0.884 | 0.569 | **0.407** |
| | NWD | 0.464 | 0.727 | 0.537 | 0.383 |
| | $\alpha$-IoU | 0.473 | 0.806 | 0.534 | 0.364 |
| | EIoU | 0.460 | 0.848 | 0.552 | 0.386 |
| | SIoU | 0.493 | 0.759 | 0.551 | 0.371 |
| | WIoU | 0.458 | 0.870 | 0.561 | 0.394 |
| | MDPIoU | 0.452 | 0.871 | 0.571 | 0.393 |
| | Shape-IoU | 0.479 | **0.897** | 0.575 | 0.387 |
| | Powerful-IoU | 0.474 | 0.855 | 0.579 | 0.385 |
| | **Focaler-IoU** | 0.476 | 0.845 | **0.582** | 0.382 |

Table 5. Ablation study on regression losses. The Focaler-IoU in PK-YOLO is replaced by CIoU, GIoU, NWD, $\alpha$-IoU, EIoU, SIoU, WIoU, MDPIoU, Shape-IoU, and Powerful-IoU, respectively. The best results are shown in bold.

## Acknowledgement

This work was supported by the Monash University Malaysia and the Ministry of Higher Education, Malaysia under Fundamental Research Grant Scheme FRGS/1/2023/ICT02/MUSM/02/1.

## References

[1] Hangbo Bao, Li Dong, Songhao Piao, and Furu Wei. Beit: Bert pre-training of image transformers. In *ICLR*, 2022. 4

[2] Giuseppe Barbato and Tommmaso Menga. Brain tumor detection. GitHub, 2022. https://github.com/giuseppebrb/BrainTumorDetection. 3

[3] Nicolas Carion, Francisco Massa, Gabriel Synnaeve, Nicolas Usunier, Alexander Kirillov, and Sergey Zagoruyko. End-to-end object detection with transformers. In *ECCV*, pages 213–229, 2020. 2

[4] Qiang Chen, Xiangbo Su, Xinyu Zhang, Jian Wang, Jiahui Chen, Yunpeng Shen, Chuchu Han, Ziliang Chen, Weixiang Xu, Fanrong Li, Shan Zhang, Kun Yao, Errui Ding, Gang Zhang, and Jingdong Wang. Lw-detr: A transformer replacement to yolo for real-time detection. arXiv preprint, arXiv:2406.03459 [cs.CV], 2024. 3, 7

[5] Qiang Chen, Jian Wang, Chuchu Han, Shan Zhang, Zexian Li, Xiaokang Chen, Jiahui Chen, Xiaodi Wang, Shuming Han, Gang Zhang, Haocheng Feng, Kun Yao, Junyu Han, Errui Ding, and Jingdong Wang. Group detr v2: Strong object detector with encoder-decoder pretraining. arXiv preprint, arXiv:2211.03594 [cs.CV], 2022. 3

[6] Xiangxiang Chu, Liang Li, and Bo Zhang. Make repvgg greater again: A quantization-aware approach. *AAAI*, 38(10):11624–11632, 2024. 4

[7] Zhigang Dai, Bolun Cai, Yugeng Lin, and Junying Chen. Up-detr: Unsupervised pre-training for object detection with transformers. In *CVPR*, pages 1601–1610, 2021. 3, 7

[8] Zhigang Dai, Bolun Cai, Yugeng Lin, and Junying Chen. Unsupervised pre-training for detection transformers. *TPAMI*, 45(11):12772–12782, 2023. 3, 7

[9] Pushkar Mahendra Desai, Rahul Vijaykumar Shabadi, Suresh Chengode, and Nasser Al-Kemyani. Multiplane imaging: A quick way to assess prosthetic aortic valve. *Annals of Cardiac Anaesthesia*, 25(2):202–203, 2022. 2

[10] Jacob Devlin, Ming-Wei Chang, Kenton Lee, and Kristina Toutanova. Bert: Pre-training of deep bidirectional transformers for language understanding. In *NACCL*, pages 4171–4186, 2019. 4

[11] Xiaohan Ding, Xiangyu Zhang, Ningning Ma, Jungong Han, Guiguang Ding, and Jian Sun. Repvgg: Making vgg-style convnets great again. In *CVPR*, pages 13728–13737, 2021. 4

[12] Alexey Dosovitskiy, Lucas Beyer, Alexander Kolesnikov, Dirk Weissenborn, Xiaohua Zhai, Thomas Unterthiner, Mostafa Dehghani, Matthias Minderer, Georg Heigold, Sylvain Gelly, Jakob Uszkoreit, and Neil Houlsby. An image is worth 16x16 words: Transformers for image recognition at scale. In *ICLR*, 2021. 3

[13] Zhora Gevorgyan. Siou loss: More powerful learning for bounding box regression. arXiv preprint, arXiv:2205.12740 [cs.CV], 2022. 8

[14] Ahmed Hamada. Br35h :: Brain tumor detection 2020. Kaggle, 2020. https://www.kaggle.com/datasets/ahmedhamada0/brain-tumor-detection. 6

[15] Jiabo He, Sarah Erfani, Xingjun Ma, James Bailey, Ying Chi, and Xian-Sheng Hua. $\alpha$-iou: A family of power intersection over union losses for bounding box regression. In *NeurIPS*, pages 20230–20242, 2021. 8

[16] Kaiming He, Xinlei Chen, Saining Xie, Yanghao Li, Piotr Dolla´r, and Ross Girshick. Masked autoencoders are scalable vision learners. In *CVPR*, pages 15979–15988, 2022. 4

[17] Kaiming He, Xiangyu Zhang, Shaoqing Ren, and Jian Sun. Spatial pyramid pooling in deep convolutional networks for visual recognition. *TPAMI*, 37(9):1904–1916, 2015. 2

[18] Priyanto Hidayatullah and Refdinal Tubagus. Yolov9 architecture explained. Stunning Vision AI, 2024. https://article.stunningvisionai.com/yolov9-architecture. 3

[19] Xiuquan Hou, Meiqin Liu, Senlin Zhang, Ping Wei, and Badong Chen. Salience detr: Enhancing detection transformer with hierarchical salience filtering refinement. In *CVPR*, 2024. 17574–17583. 3, 6, 7

[20] Jie Hu, Li Shen, and Gang Sun. Squeeze-and-excitation networks. In *CVPR*, pages 7132–7141, 2018. 4

[21] Glenn Jocher. Yolo by ultralytics (version 5.7.0). GitHub, 2022. https://github.com/ultralytics/yolov5. 2, 3, 7

[22] Glenn Jocher, Ayush Chaurasia, and Jing Qiu. rtdetr-x. GitHub, 2023. https://github.com/ultralytics/ultralytics/tree/main/ultralytics/cfg/models/rt-detr/rtdetr-x.yaml. 7

[23] Glenn Jocher, Ayush Chaurasia, and Jing Qiu. Yolo by ultralytics (version 8.1.0). GitHub, 2023. https://github.com/ultralytics/ultralytics. 2, 7

[24] Takahide Kakigi, Ryo Sakamoto, Hiroshi Tagawa, Shinichi Kuriyama, Yoshihito Goto, Masahito Nambu, Hajime Sagawa, Hitomi Numamoto, Kanae Kawai Miyake, Tsuneo Saga, Shuichi Matsuda, and Yuji Nakamoto. Diagnostic advantage of thin slice 2d mri and multiplanar reconstruction of the knee joint using deep learning based denoising approach. *Scientific Reports*, 12:10362, 2022. 2

[25] Ming Kang, Chee-Ming Ting, Fung Fung Ting, and Raphae¨l C.-W. Phan. Rcs-yolo: A fast and high-accuracy object detector for brain tumor detection. In *MICCAI*, pages 600–610, 2023. 3, 7

[26] Ming Kang, Chee-Ming Ting, Fung Fung Ting, and Raphae¨l C.-W. Phan. Bgf-yolo: Enhanced yolov8 with multiscale attentional feature fusion for brain tumor detection. In *MICCAI*, pages 35–45, 2024. 3, 7

[27] Tingting Liang, Xiaojie Chu, Yudong Liu, Yongtao Wang, Zhi Tang, Wei Chu, Jingdong Chen, and Haibin Ling. Cbnet: A composite backbone network architecture for object detection. *IEEE TIP*, 31:6893–6906, 2022. 2

[28] Tsung-Yi Lin, Michael Maire, Serge Belongie, James Hays, Pietro Perona, Deva Ramanan, Piotr Dolla´r, and C. Lawrence Zitnick. Microsoft coco: Common objects in context. In *ECCV*, pages 740–755, 2014. 2

[29] Yutong Lin, Yuhui Yuan, Zheng Zhang, Chen Li, Nanning Zheng, and Han Hu. Detr does not need multi-scale or locality design. In *ICCV*, pages 6545–6554, 2023. 3, 6

[30] Zhihao Lin, Yongtao Wang, Jinhe Zhang, and Xiaojie Chu. Dynamicdet: A unified dynamic architecture for object detection. In *CVPR*, pages 6282–6291, 2023. 2

[31] Can Liu, Kaige Wang, Qing Li, Fazhan Zhao, Kun Zhao, and Hongtu Ma. Powerful-iou: More straightforward and faster bounding box regression loss with a nonmonotonic focusing mechanism. *Neural Networks*, 170:276–284, 2024. 8

[32] Fei Liu, Huabin Wang, Shiuan-Ni Liang, Zhe Jin, Shicheng Wei, Xuejun Li, and Alzheimer's Disease Neuroimaging Initiative. Mps-ffa: A multiplane and multiscale feature fusion attention network for alzheimer's disease prediction with structural mri. *Computers in Biology and Medicine*, 157:106790, 2023. 3

[33] Yudong Liu, Yongtao Wang, Siwei Wang, Tingting Liang, Qijie Zhao, Zhi Tang, and Haibin Ling. Cbnet: A novel composite backbone network architecture for object detection. *AAAI*, 34(07):11653–11660, 2020. 2

[34] Siliang Ma and Yong Xu. Mpdiou: A loss for efficient and accurate bounding box regression. arXiv preprint, arXiv:2307.07662 [cs.CV], 2023. 8

[35] Radiological Society of North America. Brain tumor ai challenge 2021. RSNA, 2021. https://www.rsna.org/rsnai/ai-image-challenge/brain-tumor-ai-challenge-2021. 6

[36] Shoichiro Otake, Toshiaki Taoka, Masayuki Maeda, and William TC Yuh. A guide to identification and selection of axial planes in magnetic resonance imaging of the brain. *Neuroradiology Journal*, 31(4):336–344, 2018. 2

[37] PaddlePaddle. Pp-hgnet series. GitHub, 2022. https://github.com/PaddlePaddle/PaddleClas/blob/release/2.5/docs/en/models/PP-HGNet_en.md. 7

[38] PaddlePaddle. Hgnetv2. GitHub, 2023. https://github.com/PaddlePaddle/PaddleDetection/blob/develop/ppdet/modeling/backbones/hgnet_v2.py. 7

[39] Gabriele Piantadosi, Mario Sansone, Roberta Fusco, and Carlo Sansone. Multi-planar 3d breast segmentation in mri via deep convolutional neural networks. *Artificial Intelligence in Medicine*, 103:101781, 2020. 3

[40] Hamid Rezatofighi, Nathan Tsoi, JunYoung Gwak, Amir Sadeghian, Ian Reid, and Silvio Savarese. Generalized intersection over union: A metric and a loss for bounding box regression. In *CVPR*, pages 658–666, 2019. 8

[41] David Roberts. Brain tumor object detection datasets. Kaggle, 2021. https://www.kaggle.com/datasets/davidbroberts/brain-tumor-object-detection-datasets. 1, 3, 6

[42] Keyu Tian, Yi Jiang, Qishuai Diao, Chen Lin, Liwei Wang, and Zehuan Yuan. Designing bert for convolutional networks: Sparse and hierarchical masked modeling. In *ICLR*, 2023. 2

[43] Zanjia Tong, Yuhang Chen, Zewei Xu, and Rong Yu. Wise-iou: Bounding box regression loss with dynamic focusing mechanism. arXiv preprint, arXiv:2301.10051 [cs.CV], 2023. 8

[44] Ashish Vaswani, Noam Shazeer, Niki Parmar, Jakob Uszkoreit, Llion Jones, Aidan N. Gomez, Łukasz Kaiser, and Illia Polosukhin. Attention is all you need. In *NeurIPS*, pages 6000–6010, 2017. 3

[45] Ao Wang, Hui Chen, Zijia Lin, Jungong Han, and Guiguang Ding. Repvit: Revisiting mobile cnn from vit perspective. In *CVPR*, pages 15909–15920, 2024. 2

[46] Ao Wang, Hui Chen, Lihao Liu, Kai Chen, Zijia Lin, Jungong Han, and Guiguang Ding. Yolov10: Real-time end-to-end object detection. In *NeurIPS*, pages 107984–108011, 2024. 2, 7

[47] Chien-Yao Wang, Alexey Bochkovskiy, and Hong-Yuan Mark Liao. Yolov7: Trainable bag-of-freebies sets new state-of-the-art for real-time object detectors. In *CVPR*, pages 7464–7475, 2023. 2

[48] Chien-Yao Wang, Hong-Yuan Mark Liao, Yueh-Hua Wu, Ping-Yang Chen, Jun-Wei Hsieh, and I-Hau Yeh. Cspnet: A new backbone that can enhance learning capability of cnn. In *CVPRW*, pages 1571–1580, 2020. 2

[49] Chien-Yao Wang, Hong-Yuan Mark Liao, and I-Hau Yeh. Designing network design strategies through gradient path analysis. *Journal of Information Science and Engineering*, 39(4):975–995, 2023. 2

[50] Chien-Yao Wang, I-Hau Yeh, and Hong-Yuan Mark Liao. Yolov9: Learning what you want to learn using programmable gradient information. In *ECCV*, pages 1–21, 2024. 2, 7

[51] Jinwang Wang, Chang Xu, Wen Yang, and Lei Yu. A normalized gaussian wasserstein distance for tiny object detection. arXiv preprint, arXiv:2110.13389 [cs.CV], 2021. 8

[52] Zeyu Wang, Chen Li, Huiying Xu, Xinzhong Zhu, and Hongbo Li. Mamba yolo: A simple baseline for object detection with state space model. *AAAI*, 39(8):8205–8213, 2025. 2, 7

[53] Tete Xiao, Mannat Singh, Eric Mintun, Trevor Darrell, Piotr Dolla´r, and Ross Girshick. Early convolutions help transformers see better. In *NeurIPS*, pages 30392–30400, 2021. 4

[54] Jianwei Yang, Chunyuan Li, Xiyang Dai, and Jianfeng Gao. Focal modulation networks. In *NeurIPS*, pages 4203–4217, 2022. 6

[55] Hao Zhang and Shuaijie Zhang. Shape-iou: More accurate metric considering bounding box shape and scale. arXiv preprint, arXiv:2312.17663 [cs.CV], 2023. 8

[56] Hao Zhang and Shuaijie Zhang. Focaler-iou: More focused intersection over union loss. arXiv preprint, arXiv:2401.10525 [cs.CV], 2024. 2, 8

[57] Yi-Fan Zhang, Weiqiang Ren, Zhang Zhang, Zhen Jia, Liang Wang, and Tieniu Tan. Focal and efficient iou loss for accurate bounding box regression. *Neurocomputing*, 506:146–157, 2022. 8

[58] Yian Zhao, Wenyu Lv, Shangliang Xu, Jinman Wei, Guanzhong Wang, Qingqing Dang, Yi Liu, and Jie Chen. Detrs beat yolos on real-time object detection. In *CVPR*, pages 16965–16974, 2024. 3, 7

[59] Zhaohui Zheng, Ping Wang, Wei Liu, Jinze Li, Rongguang Ye, and Dongwei Ren. Distance-iou loss: Faster and better learning for bounding box regression. *AAAI*, 34(07):12993–13000, 2020. 2, 7, 8

[60] Zhuofan Zong, Guanglu Song, and Yu Liu. Detrs with collaborative hybrid assignments training. In *ICCV*, pages 6725–6735, 2023. 3, 7